\documentclass[10pt,journal,compsoc]{IEEEtran}

\usepackage[nocompress]{cite}
\usepackage{amsmath,amsfonts}
\usepackage{array}
\usepackage{listings}
\usepackage{color}
\usepackage{graphicx}
\usepackage{url}
\newcommand{\acc}{ac}
\newcommand{\est}{\widetilde{{ac}}}
\newcommand{\ess}{\widetilde{{fac}}}

\newcommand{\expect}{\mathbb{E}}

\DeclareMathOperator*{\argmax}{argmax\;}

\begin{document}

\title{Nested cross-validation when selecting classifiers is overzealous for most practical applications}

\author{Jacques Wainer, Gavin Cawley%
 \IEEEcompsocitemizethanks{\IEEEcompsocthanksitem  Computing Institute
 University of Campinas
 Campinas, SP, 13083-852, Brazil and School of Computing Sciences
 University of East Anglia
 Norwich, NR4 7TJ, U.K.\protect\\
e-mail: wainer@ic.unicamp.br and G.Cawley@uea.ac.uk}}


\maketitle

\begin{abstract}
When selecting a classification algorithm to be applied to a particular problem, one has to simultaneously select the best algorithm for that dataset \emph{and} the best set of hyperparameters for the chosen model.  The usual approach is to apply a nested cross-validation procedure; hyperparameter selection is performed in the inner cross-validation, while the outer cross-validation computes an unbiased estimate of the expected accuracy of the algorithm \emph{with cross-validation based hyperparameter tuning}.  The alternative approach, which we shall call ``flat cross-validation'', uses a single cross-validation step both to select the optimal hyperparameter values and to provide an estimate of the expected accuracy of the algorithm, that while biased may nevertheless still be used to select the best learning algorithm.  We tested both procedures using 12 different algorithms on 115 real life binary datasets and conclude that using the less computationally expensive flat cross-validation procedure will generally result in the selection of an algorithm that is, for all practical purposes, of similar quality to that selected via nested cross-validation, provided the learning algorithms have relatively few hyperparameters to be optimised.
\end{abstract}

\begin{IEEEkeywords}
Hyperparameters; classification; cross-validation; nested cross-validation; model selection 
\end{IEEEkeywords}

\section{Introduction}

A practitioner who builds a classification model has to select the best algorithm for that particular problem. There are hundreds of classification algorithms  described in the literature, such as k-nearest neighbour \cite{Dasarathy1991a}, SVM \cite{Cortes1995a}, neural networks \cite{Bishop1995a}, na\"{\i}ve Bayes \cite{Hand2001a}, gradient boosting machines \cite{friedman2001greedy}, and so on.  Although there are sometimes theoretical and/or empirical reasons to prefer a particular algorithm over another when tackling a particular problem, our current understanding of machine learning does not allow us to predict \emph{a-priori} whether one algorithm will perform better than another.   Furthermore, the so called ``no-free-lunch'' theorems even suggest that no algorithm can outperform all others for all problems \cite{Wolpert1996a}.  Therefore, for most difficult tasks, one should benefit from trying many competing algorithms to discover which gives the best performance.  However, most algorithms have one or more hyperparameters that must be set externally, for example, the k-nearest neighbour method has (usually) one hyperparameter, $k$, whereas random forest has at least two, the number of trees to be constructed and the number of features considered at each split.  Unfortunately selecting an algorithm and tuning its hyperparameters are dependent steps: an algorithm may perform very well for a  problem when using a particular set of hyperparameters, but may perform worse than other algorithms with a different, sub-optimal, set of hyperparameters.  One will therefore want to chose both the algorithm and its hyperparameters in such a way as to maximize its expected performance on future data.

Choosing an appropriate classifier (model) and optimising the hyperparameters are most often performed by minimising a cross-validation \cite{Stone} estimate of generalisation performance.  The most basic form of cross-validation, known as \emph{k-fold cross-validation} partitions the available data into $k$ disjoint chunks of approximately equal size.  In each iteration a training set is formed from a different combination of $k-1$ chunks, with the remaining chunk used as the test set; a model is then fit to the training set and its performance evaluated using the test set.  The average of the performance metric on the test set in each iteration is then used as an estimate of the generalisation performance of a model fit to all of the available data.  There are two common procedures for selecting the best algorithm and tuning the hyperparameters via cross-validation, the first is called \emph{nested cross-validation}, also known as \emph{double cross-validation} \cite{Stone}, the second appears to have no standard name, so we will call it \emph{flat cross-validation}:
\begin{LaTeXdescription}
\item[Flat Cross-Validation:] The hyperparameters of each model are tuned to minimise a cross-validation based
                                 estimate of generalisation performance.  The cross-validation performance estimate,
                                 evaluated for those optimal hyperparameter values, is then used to select the best
                                 model to use in operation.  This approach is computationally inexpensive, however
                                 an optimistic bias is introduced into the performance estimate as it has been 
                                 directly optimised in tuning the hyperparameters \cite{Cawley2010}.  Unless this
                                 bias is commensurate for all of the candidate models, the re-use of the 
                                 hyperparameter optimisation criterion as a model selection criterion may result in
                                 a sub-optimal choice of model, potentially selecting a model that is particularly
                                 susceptible to this bias, rather than a model with genuinely higher performance.
\item[Nested Cross-Validation:] An outer cross-validation procedure is performed to provide a performance estimate 
                                   used to select the optimal model.  In each fold of the outer cross-validation, the
                                   hyperparameters of the model are tuned independently to minimise an inner 
                                   cross-validation estimate of generalisation performance.  The outer 
                                   cross-validation is then essentially estimating the performance of a method for 
                                   fitting a model, cross-validation based hyperparameter tuning.  This eliminates 
                                   the bias introduced by the flat cross-validation procedure as the test data in 
                                   each iteration of the outer cross-validation has not been used to optimise the 
                                   performance of the model in any way, and may therefore provide a more reliable
                                   criterion for choosing the best model.  The computational expense of nested
                                   cross-validation, however, is substantially higher.
   
\end{LaTeXdescription}
The aim of this study is to perform an empirical evaluation to determine whether the additional computational 
expense of the nested cross-validation procedure is generally justified by providing a more reliable means of choosing the best model and statistically superior performance.  While no empirical study can possibly cover \emph{every} learning task or apply to \emph{all} learning systems, a thorough study using a large suite of benchmark datasets and a set of learning systems representative of those used in applied work remains the best available guidance for practitioners.

\subsection{Estimating the Generalisation Performance of a Model}

Let us denote by $\acc(Y|a,X, \theta_a)$, the accuracy of algorithm $a$  when trained on data $X$ with hyperparameters $\theta_a$ and tested on data $Y$.  Let us assume that a data set $G$ is an i.i.d.\ sample from some underlying distribution $\mathcal{D}$. The best algorithm for the data set $G$ is the algorithm that when trained on the whole data set $G$, with the optimal values for the hyperparameters, will have the highest expected accuracy for future data. 

The expected accuracy for future data (for algorithm $a$ trained on $G$ with hyperparameters $\theta_a$ is:
\[ \expect_{g \sim \mathcal{D}} [\acc(g|a,G,\theta_{a})].
\]
Given a set of candidate classification algorithms, $\mathcal{A}$, the best algorithm $a_{best}$ is then:
\begin{equation}\label{eq:1}
   a_{best} = \argmax_{a \in \mathcal{A}} \;  \expect_{g \sim \mathcal{D}} [\acc(g|a,G,\hat{\theta}_{a})],
\end{equation}
where $\hat{\theta}_a $  denote the best set of hyperparameters for algorithm $a$, that is: 
\begin{equation}\label{eq:2}
   \hat{\theta}_a = \argmax_{\theta_a} \; \expect_{g \sim \mathcal{D}} [\acc(g|a,G,\theta_{a})].
\end{equation}

Both nested and flat cross-validation procedures estimate the expected performance of the classifier, with optimal hyperparameter settings,
\begin{equation}\label{eq:term} 
   \expect_{g \sim \mathcal{D}} [\acc(g|a,G,\hat{\theta}_{a})],
\end{equation}
and then select the algorithm, $a \in \mathcal{A}$, having the highest estimate. Let us denote as $\est^{f}(a,G)$ the flat cross-validation estimate of the term in Equation~\ref{eq:term}, and  $\est^{n}(a,G)$, the nested cross-validation estimate.  Both estimates, $\est^{f}(a,G)$ and $\est^{n}(a,G)$, will likely result in different numeric values (Section~\ref{sec:cross}). 

The nested cross-validation (CV) procedure is considered the more appropriate because $\est^{n}(a,G)$ is an unbiased estimate of the expectation in Equation~\ref{eq:term}, whereas $\est^{f}(a,G)$ has a positive bias \cite{Cawley2010}, that is, on average, $\est^{f}(a,G)$ will have higher, overly optimistic values than $\expect_{g \sim \mathcal{D}} [\acc(g|a,G,\hat{\theta}_{a})]$.  This arises as the data used in performance evaluation are also used indirectly in tuning the hyperparameters.  Thus the algorithm selected by the nested procedure is considered the more ``correct'' because the estimates of that procedure are unbiased relative to the true expected accuracy. However, the nested procedure has a much higher computational cost than the flat procedure. Therefore one may want to use the flat procedure, even at some risk of not selecting the best algorithm, where the computational expense is prohibitive. Notice that for the purpose of algorithm selection the positive bias of the flat procedure is not itself a problem provided the algorithm ranked highest by the nested procedure is the same as that highest ranked by the flat procedure, which implies the degree of bias is approximately the same for all classifiers.

Let us assume that $n_G$ is the algorithm selected by the nested CV procedure, and $f_G$ the algorithm selected using flat CV for data set $G$. Clearly if $n_G=f_G$ in a very high percentage of cases, then one may choose the less expensive procedure, at some slight risk. In the cases where the two procedures do not agree on the best algorithm, we will compute the accuracy gain of the nested procedure selection relative to the flat selection procedure, or in other words, the difference in the expected accuracy on future data of the nested selection and the flat selection. 

Let $\hat{\theta}_{n_G}$ be the optimal hyperparameters for the $n_G$ algorithm, and $\hat{\theta}_{f_G}$ those of the $f_G$ algorithm, as computed by the expression~\ref{eq:2}.  Let us define the accuracy gain of using the nested CV procedure on dataset $G$ as: 
\begin{equation}
  \label{eq:6}
  \mbox{accgain}(G) = \expect_{g \sim \mathcal{D}} [\acc(g|n,G,
\hat{\theta}_{n})] - \expect_{g \sim \mathcal{D}} [\acc(g|f,G,
\hat{\theta}_{f})],
\end{equation}
where we left implicit the subscript ${}_G$ for the $n$ and $f$ symbols, for the sake of clarity.  Of course, one cannot determine the true value of the accuracy gain, but as  we will discuss in Section~\ref{sec:exper-proc}, we will be able to estimate it.  

\subsection{Flat and nested CV estimates}\label{sec:cross}

Both nested and flat CV procedures rely on using cross-validation to estimate an expectation of the form $\expect_{g \sim \mathcal{D}} [\acc(g|a,G, \theta_{a})] $ when $\mathcal{D}$ is not known.  Given a set of data $G$, cross-validation defines a set of pairs of sets $TR_k$ and $TE_k$ where $TR_k$ is called a \emph{training set}, and $TE_k$ is called \emph{test set} or sometimes \emph{validation set}, and where:
\[ TR_k \cap TE_k = \emptyset \quad \mbox{and} \quad TR_k \cup TE_k  \subseteq G.\] 
Common cross-validation procedures include: k-fold, bootstrap, leave-one-out, hold-out, and so on.

Given a particular cross-validation procedure (which given $G$ defines the sets $TR_k$ and $TE_k$ and the number of such pairs), the cross-validation estimate for the expected accuracy of the classifier (for a particular algorithm $a$ and hyperparameters $\theta$) is calculated as: 
\[ \mbox{mean}_k \; \acc(TE_k|a,TR_k, \theta). \]
The flat CV estimate of $\expect_g [\acc(g|a,G, \hat{\theta}_{a})]$ will select  $\hat{\theta}^{f}_a$  as the value of $\theta$ that maximizes $\mbox{mean}_k \acc(TE_k|a,TR_k, \theta) $:
\begin{equation}\label{besthyp}
 \hat{\theta}^{f}_a = \argmax_{\theta} \; \mbox{mean}_k \; \acc(TE_k|a,TR_k,  \theta),
\end{equation}
and then use $\hat{\theta}^{f}_a$ to estimate $\expect_g [\acc(g|a,G,\hat{\theta}_{a})]$, that is:
\begin{equation}\label{flatest}
\est^{f}(a,G)  = \mbox{mean}_k \; \acc(TE_k|a,TR_k,  \hat{\theta}^{f}_a).
\end{equation}  

In  nested CV, each training set (of the outer cross-validation) $TR_k$ is further subdivided into pairs of sets of data $TR_{km}$ and $TE_{km}$ where again:  
\[ TR_{km} \cap TE_{km} = \emptyset \quad \mbox{and} \quad TR_{km} \cup TE_{km}  \subseteq TR_k.\]
The nested cross-validation procedure will select the best hyperparameter $\hat{\theta}^k_a$ for each training set $TR_k$  as: 
\[ \hat{\theta}^k_a = \argmax_{\theta} \; \mbox{mean}_j \; \acc(TE_{kj}|a,TR_{kj}, \theta ). \]
The nested CV estimate of the expected accuracy for future data is:
\[ \est^{n}(a,G)  = \mbox{mean}_k \; \acc(TE_k|a,TR_k, \hat{\theta}^k_a ).\]

Figure~\ref{lstflat} gives an implementation of the flat CV as a Python program and Figure~\ref{lstnested} provides the corresponding implementation for the nested CV procedure where the following functions are assumed: 
\begin{itemize}

   \item \texttt{createCV(data,...)} creates a list of pairs \texttt{(train,test)}   from the \texttt{data}. Other parameters may include, for  example, $k$ if a k-fold CV procedure is used, or the proportion of cases in the training set, if a hold-out procedure is used. 
   
   \item \texttt{createGrid()} creates the list of hyperparameter tuples to be tested, forming a regular grid.
   
   \item \texttt{classtrain(train,theta)} returns the  classifier trained on data \texttt{train} with hyperparameters set to \texttt{theta}.

   \item \texttt{accuracy(model,test)} returns the accuracy (or any other quality measure) for the classifier \texttt{model} when run on data \texttt{test}.
   
\end{itemize}

\begin{figure}
\small
\begin{lstlisting}[language=Python]
def flat(data,...):
  cv.lst = createCV(data,...)
  accmax=0.0
  for theta in createGrid(...):
     acc=0.0
     for train,test in cv.lst:
        model=classtrain(train,theta)
        acc=acc+accuracy(model,test)
     if acc > accmax:
        accmax=acc
        thetamax=theta
  return accmax/len(cv.lst)
\end{lstlisting}  
\caption{Implementation of the flat cross-validation procedure as a Python program.}\label{lstflat}
\end{figure}

\begin{figure}
\small
\begin{lstlisting}[language=Python]
def nested(data,...):
  accfinal=0.0
  cv.lst = createCV(data,...)
  for tr.o,te.o in cv.lst:
     accmax=0.0
     for theta in createGrid(...):
       acc=0.0
       cv.inner.lst = createCV(tr.o,...)
       for tr.i,te.i in cv.inner.lst:
          model=classtrain(tr.i,theta)
          acc=acc+accuracy(model,te.i)
       if acc > accmax:
          accmax = acc
          thetamax = theta
     model2 = classtrain(tr.o,thetamax)
     accfinal = accfinal+accuracy(model2,te.o)   
  return accfinal/len(cv.list)
\end{lstlisting}
\caption{Implementation of the nested cross-validation procedure as a Python program.}\label{lstnested}
\end{figure}

Note that the nested CV procedure does not calculate a single best set of hyperparameter values; each training set of the outer cross-validation ($k$) may select a different set of ``optimal'' hyperparameters.  In practice, the hyper-parameter values of the classifier used in operation are found by cross-validation using all available data.  These
are, of course, the hyperparameter values determined by the flat cross-validation procedure.
%
%
Essentially nested cross-validation estimates the performance of the full procedure used to generate the final model, including hyperparameter tuning.  This provides an unbiased estimator for choosing the best classifier system, but does not affect the operational hyperparameter values.

The research presented here evaluated the mean accuracy gain of the nested CV procedure over flat-CV, by estimating its value over 115 real-life datasets, for 12 different classification algorithms.  We show that the expected accuracy gain is very small, and we argue that the gain is of negligible practical consequence for most applications.  That is, in the majority of cases, either the selection of the flat and nested procedures coincide, or where they differ the algorithms selected by each approach are so close in terms of expected accuracy that this difference can be considered irrelevant, provided the algorithms have relatively few tunable hyperparameters (as this strongly influences the bias of the flat-CV procedure).

\section{Data and Methods}
\label{sec:data-methods}

\subsection{Experimental procedure}
\label{sec:exper-proc}

In this section, we set out in general terms the experimental procedure followed by this research.  We performed 6 repetitions of a 50\% split of each data set into train and test subsets, each with the same proportion of patterns belonging to each class. For each dataset $D_i$, $TR^i_r$ is the training subset for repetition $r$ and $TE^i_r$ is the corresponding test subset.  For each train set,  $TR^i_r$, we computed the  expected accuracy using a 5-fold-within-5-fold nested-CV procedure ($\est^{n}(a,i,r)$) and using a 5-fold flat-CV procedure ($\est^{f}(a,i,r)$) for 12 different classification algorithms $a$ (the classification algorithms are discussed in section~\ref{sec:datasets}).  The flat-CV procedure also determines the best selection of hyperparameters ($\hat{\theta}_{air}$) for each algorithm $a$, for each $TR^i_r$. \footnote{Following the nested cross-validation procedure, the selected model is re-trained on all of the available data, with 5-fold cross-validation based tuning of the hyperparameter values, which will of course give the same hyperparameter values as those already determined from the flat cross-validation trials.}

Let us define $f(i,r)$ as the algorithm selected by the flat procedure on $TR^i_r$, and $n(i,r)$ as the algorithm selected by the nested procedure.  The future accuracy of an algorithm $a$ on repetition $r$ for data set $i$ is the accuracy of the algorithm when trained on $TR^i_r$ with the best hyperparameters selected by the flat procedure $\hat{\theta}_{air}$ and tested on $TE^i_r$. In particular we are interested in the future accuracy of the algorithms selected by the nested procedure $n(i,r)$ and by the flat procedure $f(i,r)$, and will define $\ess^n(i,r)$ as the future accuracy of the nested selection (for data set $i$ and round $r$) -- similarly $\ess^f(i,r)$ is the future accuracy of the flat selection. Formally:
\begin{align} \label{eq10}
\begin{split}
\ess^n(i,r) &= \acc(TE^i_r|n(i,r),TR^i_r,  \hat{\theta}_{air}), \\
\ess^f(i,r) &= \acc(TE^i_r|f(i,r),TR^i_r,  \hat{\theta}_{air}). 
\end{split}
\end{align}

\noindent The accuracy gain of using the nested procedure instead of the flat procedure is the difference between the future accuracy of the  nested selection and the future accuracy of the flat selection,
\begin{align} \label{eq11} 
accgain(i,r) &= \ess^n(i,r)-\ess^f(i,r). 
\end{align}

\noindent Finally, the accuracy gain of a data set $i$ is the average of the accuracy gains for the six rounds for that data set:

\begin{align} \label{eq12} 
accgain(i) &= \frac{1}{6}\sum_{r=1}^6 accgain(i,r).
\end{align}

Since the nested procedure is considered the ``more correct'' one, it should select the ``more correct''  algorithm, and thus it is more likely that the future accuracy of the nested selection would be higher than that of the flat selection. Thus, in general one would expect a positive accuracy gain.

To show that the least costly flat procedure achieves similar results
(in future accuracy) as the nested procedure, we must show that the
accuracy gains over all data sets is small. Unfortunately there is no
standard way of showing that an ``aggregated'' accuracy gain is
small. A null hypothesis test will only determine if the aggregated
accuracy gain is significantly different than 0, but a) even if it is
significantly different than 0 that difference may not be sufficiently
large to be of \emph{practical} significance, and b) if the accuracy
gain is not significantly different than 0 that does not establish
that it actually is small, unless the statistical power of the test is
high.

Thus, to provide evidence that the accuracy gains are small, we define
a threshold of irrelevance for each data set $\delta(i)$, which is the
change in accuracy one should consider as irrelevant or of no
``practical significance''. Below we discuss our proposal for this
threshold. Given $\delta(i)$ we want to show that:

\begin{equation} \label{measure} 
\mbox{~in general~} \quad |accgain(i)| < \delta(i).
\end{equation}

We use the Wilcoxon signed rank test (a paired non-parametric test) to
show that the median of the set $\{\: |accgain(i)|\: \}$ for all data
sets $i$ is smaller and significantly different than the median of the
set $\{ \: \delta(i) \: \}$ We also report the mean and the 95\%
confidence interval of
$\{ |accgain(i)| - \delta(i) | \mbox{~for all~} i \}$ so the reader
may gain a sense of the magnitude of the differences. The confidence
interval was calculated using bootstrap with 5000 rounds.

The idea for a threshold of irrelevance is based on unavoidable errors
in the accuracy estimate; unavoidable because they depend on random
factors, such as the sampling of the data to form training and test
sets. The threshold depends both on the data set and the algorithm. If
the data set is small one expects larger changes in accuracy when
different splits of train and test or when comparing estimated
accuracy with the real accuracy on future, unseen data. If the
algorithms overfits the data, or if the algorithm underfits the data,
one would also expect larger differences in the accuracy in those
different conditions.

Our proposal for the irrelevance threshold $\delta$ is based on the
idea that the nested procedure estimate of the future accuracy is only
an estimate of the actual generalisation performance.  Differences
between the estimate and the measured accuracy for some unseen data
may indicate how sensitive is the combination of data set and
algorithm to these unavoidable variations.  We define $\Delta(a,i,r)$
as the difference between the nested estimate of future accuracy and
the measured future accuracy for a particular algorithm $a$, data set
$i$ an repetition $r$. Formally:
\begin{equation}\label{delta1}
\Delta(a,i,r) = | \est^n(a,i,r) - \ess(a,i,r) |.
\end{equation}

The threshold of irrelevance for a data set $i$ and round $r$ ,
$\delta(i,r)$, is the minimum between  $\Delta(n(i,r),i,r)$ and
$\Delta(f(i,r),i,r)$ 
\begin{align} \label{eq13x} 
\delta(i,r) &= \min \Delta(n(i,r),i,r), \Delta(f(i,r),i,r),
\end{align} 
The idea is that the threshold of irrelevance for a data set and a
round is the smallest of the errors between estimated and measured
future accuracy for the two ``important/best''  algoritms for that
data set and for that round, $n(i,r)$ and $f(i,r)$. The reason to take
the minimum is to achieve a more restrictive definition of
irrelevance. 

The final threshold for the data set $\delta(i)$ is the average of
$\delta(i,r)$ for all repetitions:
\begin{align} \label{eq13} 
\delta(i) &= \frac{1}{6} \sum_r  \delta(i,r)
\end{align}

Finally, it is interesting to understand the role of the repetition in
this experimental procedure. Repetitions $r$ are seen as different
experiments to compute the accuracy gain of the nested procedure
versus the flat procedure.  Each repetition may select different
algorithms in the nested and in the flat procedures. The goal of the
experiment/repetition is to compute the accuracy gain
(Equation~\ref{eq11}) and the irrelevance threshold
(Equation~\ref{eq13x}). Only then are the accuracy gain and
irrelevance thresholds aggregated across repetitions on the same data
set (Equations~\ref{eq12} and ~\ref{eq13}).

This form of analysis is inspired by the nested cross-validation
procedure, which only aggregates the data on the different
folds/hold-out subsets to compute the final measure of interest, the
expected accuracy. The two measures of interest in this research are
the accuracy gain and the threshold of irrelevance, and only at that
level, the results are averaged across repetitions. \ref{othermet}
discusses different ways of using the repetitions and presents the
corresponding results.  \ref{othermet} also presents a different
definition of the threshold of irrelevance for a data set, and the
corresponding results.

\subsection{Scenarios}

In this paper we are interested in answering two questions regarding
nested and flat procedure. The first question is whether one need use
a nested procedure to select the best among three very good algorithms
for classification: random forest (rf), SVM with RBF kernel
(svmRadial), and gradient boosting machine (gbm). There is some
independent evidence to suggest that these three algorithms are likely
the best classification algorithms in general.  Fern\'andez-Delgado \emph{et al.} \cite{delgado14} do not
test gradient booting machines, and find that random forest and SVM
with RBF kernel are the two best families of algorithms. Wainer \cite{arxiv}
does test gradient boosting machines, and finds that those three form
the best three families of classification algorithms. As we will
discuss in section~\ref{sec:results}, this research finds that
random forest is the algorithm with lowest mean rank, followed by the SVM
with an RBF kernel, followed by gradient boosting machines. Thus, practitioners
that have a restriction on the time needed to select the best
classification algorithm should restrict themselves to these three
algorithms. The first question we will address is whether, when
selecting among rf, svmRadial, and gbm one can avoid the nested
procedure and use a flat procedure instead.  In this scenario, called
\textbf{top3}, we restrict the analysis to only those three
algorithms.

The second question is whether the nested procedure is necessary when
 a wider range of classifiers are being compared.  In this case, we tested 12
different families of classifiers (the algorithms are discussed in
Section~\ref{sec:datasets}). We call this the \textbf{full} scenario.

\subsection{Datasets and classification algorithms}
\label{sec:datasets}

We used the suite of data sets collected from the UCI public
repository and processed by Fern\'andez-Delgado \emph{et al.} \cite{delgado14} and further processed by
Wainer \cite{arxiv}, such that all data sets are binary classification tasks.
For the 9 datasets with more than 10000 data points, we applied the
procedures (nested and flat CV) on only a random subset of 5000 data
points (from each subset).  For each subset, we applied 12 different
classification algorithms.  The algorithms and their abbreviations are
as follows:
\begin{LaTeXdescription}
\item[bst] A boosting of linear classifiers.
\item[gbm] Gradient boosting machines - a boosting of short decision trees \cite{friedman2001greedy}.
\item[knn] The k-nearest neighbours classifier \cite{Dasarathy1991a}.
\item[lvq] Learning vector quantization \cite{kohonen1995learning}.
\item[nb] Naive Bayes classifier \cite{Hand2001a}.
\item[nnet] A 1-hidden layer neural network with sigmoid transfer function \cite{Bishop1995a}.
\item[rf] Random forest - a bagging of decision trees \cite{Ho1998a}.
\item[rknn] A bagging of knn classifiers on a random subset of the
  original features \cite{Dasarathy1991a}.
\item[sda] A L1 regularized linear discriminant classifier \cite{ahdesmaki2010feature}.
\item[svmLinear] A SVM with linear kernel \cite{Cortes1995a}.
\item[svmPoly] A SVM with polynomial kernel \cite{Cortes1995a}.
\item[svmRadial] A SVM with RBF kernel \cite{Cortes1995a}.
\end{LaTeXdescription}
Details of the particular implementations of these algorithms and hyperparameter search grid are described in \cite{arxiv}.

\subsection{Reproducibility}

The data sets used in the paper are available at
\url{https://figshare.com/s/d0b30e4ee58b3c9323fb} as described in
\cite{arxiv}. The program to run the different procedures and the
different classifiers, the results of the multiple runs, and the R
program to perform the statistical analysis described in this paper
are available at \url{https://figshare.com/s/85887b70d4a380736928}~.

\section{Results}
\label{sec:results}

Table~\ref{tab1} lists the mean rankings of the algorithms, according to the nested CV estimate of their accuracies, over all repetitions and over all data sets.

\begin{table}[ht]
\caption{Ranking of the algorithms based on the mean rank for each repetition.}\label{tab1}
\centering
\begin{tabular}{lr}
  \hline
 algorithm & mean rank \\ 
  \hline
   rf & 3.4 \\ 
  svmRadial & 3.6 \\ 
  gbm & 4.0 \\ 
  nnet & 4.8 \\ 
  rknn & 5.3 \\ 
  svmPoly & 5.3 \\ 
  knn & 5.4 \\ 
  svmLinear & 6.1 \\ 
  sda & 6.6 \\ 
  lvq & 6.7 \\ 
  nb & 7.9 \\ 
  bst & 8.7 \\ 
   \hline
\end{tabular}

\end{table}

\noindent The results of the top 3 agree with the order in \cite{arxiv}. 

\subsection{Results for the top 3 and full scenarios}
\label{resall}

\begin{figure*}[t]
\begin{center}
\includegraphics[width=\textwidth,height=0.5\textwidth]{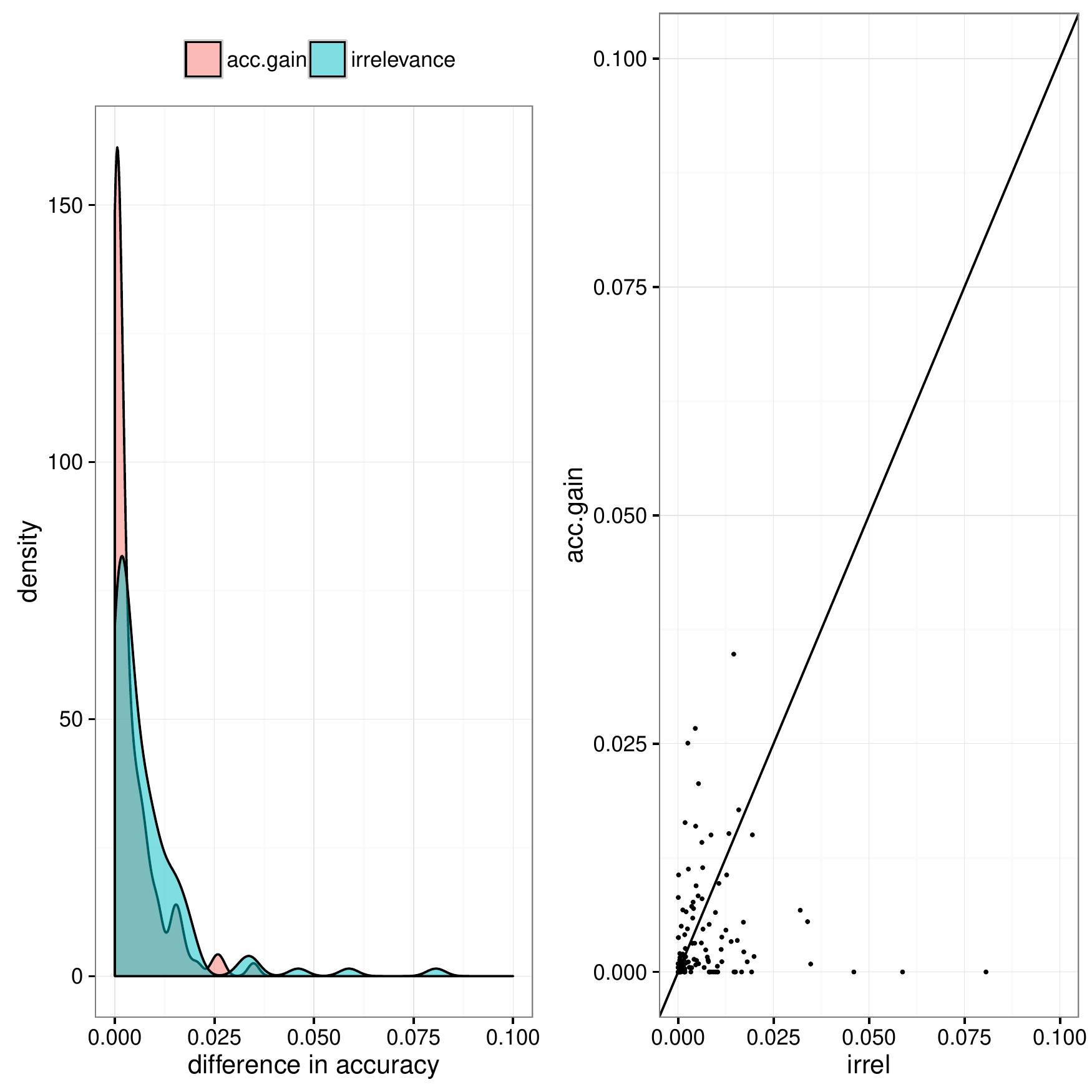}
\caption{The distribution and comparison of accuracy gain and irrelevance threshold - top 3 scenario} \label{fig3}
\end{center}
\end{figure*}

\begin{figure*}[t]
\begin{center}
\includegraphics[width=\textwidth,height=0.5\textwidth]{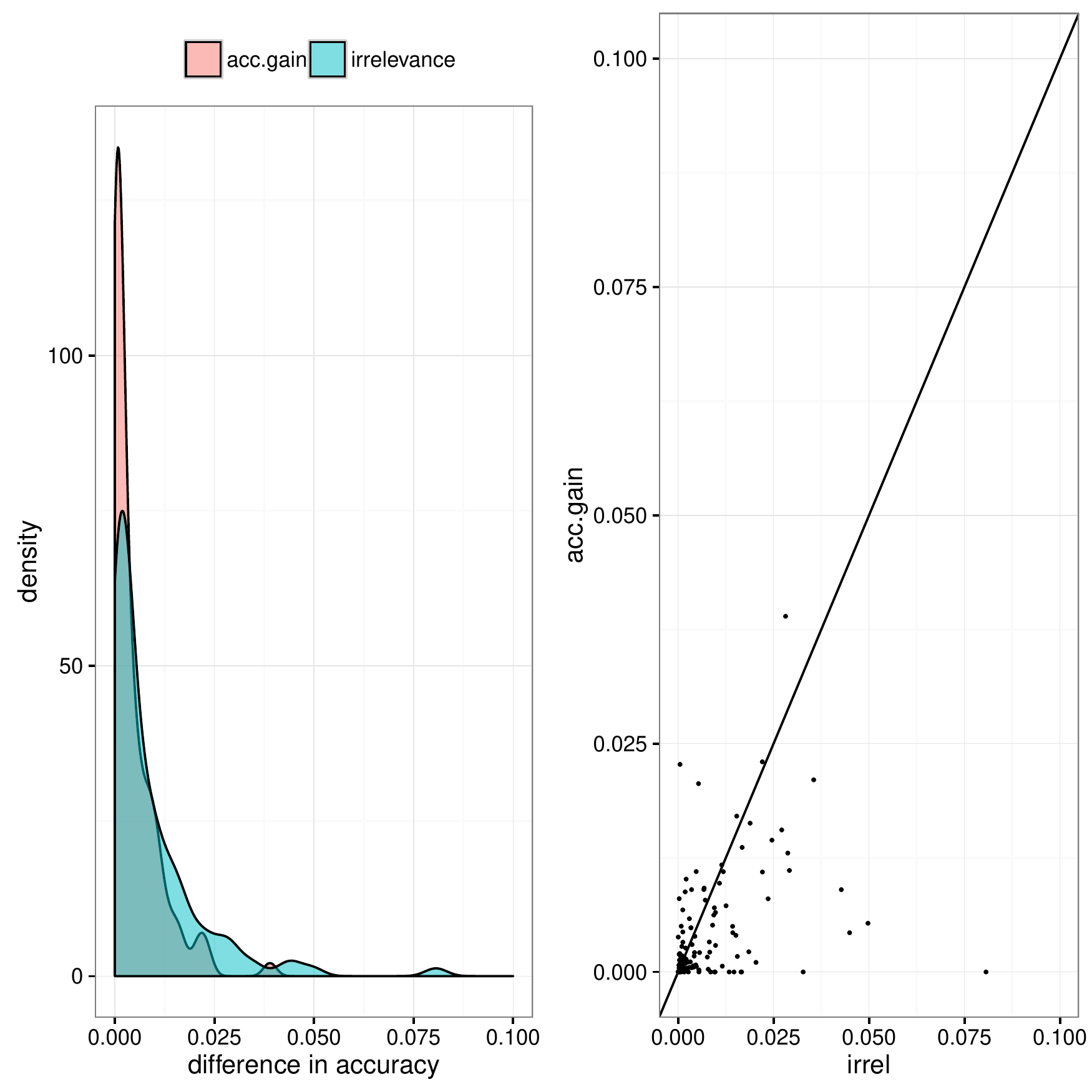}
\caption{The distribution and comparison of accuracy gain and irrelevance threshold - full scenario} \label{fig4}
\end{center}
\end{figure*}

Figure~\ref{fig3} displays the accuracy gain and the thresholds of
irrelevance for the top3 scenario (random forest, SVM with RBF
kernel, and gradient boosting machines). The first figure is the
distributions of the absolute value of the accuracy gain and the data
set thresholds of irrelevance. The second figure relates each measure
of accuracy gain (in the vertical) with the corresponding threshold of
irrelevance (horizontal). Notice that most points are in the lower
part of the $y=x$ line, which show that in most cases, the threshold
of irrelevance is higher than the corresponding accuracy gain.
Figure~\ref{fig4} displays the corresponding distributions and
comparisons of the accuracy gain and threshold of irrelevance for the
full scenario.

Table~\ref{tab3} displays the results for statistical analysis for the
top 3 and full scenarios. ``Same choice'' is the proportion of times
the algorithm selected using flat CV agreed with that selected using
nested CV.  The column ``p.value'' is the p-value of the one-sided
Wilcoxon signed rank test between the accuracy gain and the
irrelevance threshold. The ``mean'' column is the mean of the
difference of the accuracy gain and the irrelevance threshold, and it
is negative as expected, the ``low CI'' and ``high CI'' columns are
the lower and higher limits of the 95\% confidence interval for the
mean.

For the top 3 scenario the flat procedure selects the same algorithm
that the nested procedure selects in 71\% of the cases (a random
choice would give a figure of 33\%). The p-value is below 0.05, which
shows that the accuracy gains for the nested procedure are
significantly smaller than the corresponding thresholds of irrelevance
A similar conclusion can be reached by inspecting the confidence
interval of the mean, which does not include the 0. Therefore, either
using the p-value or the confidence interval one can claim that the
accuracy gain is statistically significantly less than the the
corresponding irrelevance threshold (at the 95\% level of
significance).  Thus our claim that there is no practical difference
on average between using either the nested or the flat procedure to
select among random forest, SVM with RBF kernel, and gradient boosting
machines.  For the full scenario, the agreement rate between flat and
nested is 62\% (against 8\% if the decision was random), and again a
p-value below 0.05 and the confidence interval does not include the
0. Therefore again one can be confident that on average the accuracy
gain is below the corresponding threshold of irrelevance.

\begin{table}[ht]
\caption{The results for the selection of the top 3 and full scenarios.  The column ``Same choice'' is the proportion of times the selection using flat CV agreed with the selection using nested CV. ``p.value'' is the p-value of the one sided Wilcoxon signed rank test of the accuracy gain and the corresponding threshold. ``Mean'' is the mean value of the difference  $|accgain(i)| - \delta(i)$, and ``low CI'' and ``high CI'' are the limits of the 95\% confidence interval of that mean. }\label{tab3}
\centering
\begin{tabular}{lrrrrr}
  \hline
scenario &   same choice   &   p.value & mean & low CI & high CI\\
\hline
top 3 & 71\% &  0.001 & -0.004 & -0.007 & -0.002\\
full & 62\% & 3.6e-06 &  -0.004 & -0.007 &  -0.003  \\ 
   \hline
\end{tabular}
\end{table}

\section{Discussion}
\label{sec:discussion}

This paper makes two claims:
\begin{itemize}
\item Nested CV procedures are probably not needed when selecting from
  among random forest, SVM with Gaussian kernel, and gradient boosting
  machines (which are on average the three best classification
  algorithms for the suite of data sets used in this research).
\item Nested CV procedures are probably not needed when selecting from
  among any set of classifier algorithms (provided they have only a
  limited number of hyper-parameters that must be tuned as we will
  discuss below).
\end{itemize}
The first claim was explicitly tested on 115 data sets and thus to
generalize it, the reader must believe that the 115 datasets are a
unbiased sample of future data sets a practitioner will face in the
future. We discuss the limits of such generalization below.  This
second claim caries another risk for generalization, namely that the
full set of 12 classifiers is a good sample of future sets of
classifiers that will be selected in future applications.

Wainer \cite{arxiv} discusses some of the limits to the generalization of
conclusions obtained from the set of 115 data sets to any future data
set and we briefly summarise them here.  The data sets tested in this
research were only of medium size (up to 100,000 data points), only
binary data sets were used, and none of them are derived from text
classification problems (with high dimensionality and high sparsity).
It is not immediately obvious how the number of dimensions, sparsity,
or the fact that there are more than two classes could have a
substantial impact on the claims made in this research.  Data set size
could be an issue, however, as the bias introduced by the flat
cross-validation procedure generally decreases as the size of the
dataset increases.  

Table~\ref{tablarge} reports the statistical tests
for the two scenarios, only for the 32 data sets with 2000 data or
more.  For the larger data sets only, the strength of the evidence (the p-value or the range of the CI) in
favour of the practical equivalence of the nested and flat procedure
diminishes, as expected, given that there are fewer datasets/measures
used in the significance test, but the effect size (the mean of the difference between the accuracy gain and the threshold of irrelevance) remains very similar to the ones in Table~\ref{tab3}, and the proportion of cases with the same choice for both procedures increases for the larger data sets.  We believe that one can safely make the claim of the
practical equivalence of the nested and flat procedures even for larger datasets than have been tested in this research. 

\begin{table}[ht]
\caption{The results for 32 data sets with at least 2000 data points. }\label{tablarge}
\centering
\begin{tabular}{lrrrrr}
  \hline
scenario &  same choice   & p.value& mean & low CI & high CI\\
\hline
top 3 & 80\%&0.0016 & -0.005 & -0.014 &  -0.002 \\
full & 71\% & 0.0108 & -0.004 & -0.015 & -0.001 \\
\hline 
\end{tabular}
\end{table}

The second generalization, that our analysis of all 12 classification
algorithms is a sample of any future selection choice the practitioner
will face in the future has at least one severe limit. All algorithms
tested in this research had a small number of hyperparameters, from 1
to at most 3 (for the gbm). Cawley and Talbot \cite{Cawley2010} show an interesting
example of a LS-SVM classifier with ARD kernel (automatic relevance
detection) which means that each original data dimension has its own
$\gamma$ hyperparameter (of a RBF kernel). In principle, a LS-SVM with
ARD subsumes the standard RBF LS-SVM, and thus it should not have an
expected error higher than the classical RBF LS-SVM. But
Cawley and Talbot \cite{Cawley2010} show that although the LS-SVM with ARD kernel
achieves a lower expected error when using the flat CV estimate of the
error, when using the nested procedure, the classical RBF LS-SVM has a
statistically significant lower error in 7 out of 13 data sets tested,
while the ARC LS-SVM is statistically better in only one of those 13
data sets \cite[Table 2]{Cawley2010}. In this case, because the ARD
LS-SVM has so many more hyperparameters than the RBF LS-SVM, the flat
procedure will likely overfit the data. Thus, in the case of
algorithms with very different number of hyperparameters (such as ARD
based algorithms or deep networks) we feel less confidence in the
practical equivalence results between the nested and flat procedures.

Appendix~\ref{othermet} shows that the conclusions reached by this
paper do not strongly depend on the method of analysis - two other
methods of analysis result in the same
conclusions. Appendix~\ref{othermet} also shows that the results
remain even when a different definition of the threshold of
irrelevance is used. Appendix~\ref{oneshot} shows that one should not
go a step further and skip the selection of the algorithm altogether -
in that case mean accuracy gain is significantly larger than the
threshold of irrelevance.

The results in this paper are only applicable to practitioners, that
is, for users that have the goal of selecting the likely best
classification algorithm to solve a particular problem. Our results
cannot be applied by a scientist whose goal is to provide evidence
that one classification algorithm is better than another. Our claim of practical
equivalence applies only to the best ranked algorithm for both
procedures, and not that the two procedures have some significant
agreement regarding the full ranking. For example, Table~\ref{tabx}
list the rank of the 12 algorithms when using the flat procedure
estimate to order them.  The table should be compared to
Table~\ref{tab1}. The order of the algorithms is very different; in
particular using the flat estimate, the gbm would be classified as the
best algorithm while using the nested CV estimate, it is ranked
third. In particular, given that the gbm has 1 hyperparameter more
than svmRadial or rf, we believe that this improvement in the ranking
could be due to the model overfitting described above
\cite{Cawley2010}.

\begin{table}[ht]
\caption{Ranking of the algorithms based on the mean rank for each subset ordered by the flat CV estimate of the expected error.}\label{tabx}
\centering
\begin{tabular}{lr}
  \hline
 algorithm  & mean flat-CV rank \\ 
 \hline
  gbm & 3.0 \\ 
  svmRadial & 3.2 \\ 
  rf & 4.0 \\ 
  nnet & 4.1 \\ 
  rknn & 4.2 \\ 
  svmPoly & 5.2 \\ 
  knn & 5.3 \\ 
  lvq & 6.4 \\ 
  svmLinear & 6.4 \\ 
  sda & 7.0 \\ 
  nb & 8.4 \\ 
  bst & 8.6 \\ 
   \hline
\end{tabular}
\end{table}

\section{Conclusion}
\label{sec:conclusion}

There is very strong evidence that when selecting among a random
forest, a SVM with Gaussian kernel, and a gradient boosting machine
(the three best algorithms on average for the 115 real life datasets
tested) one can generally use the flat cross-validation procedure to
both search for the best hyperparameters and to select the best
algorithm itself. Our analysis shows that the algorithm selected by
the flat procedure will, on average, perform as well as the one that
would be selected by the nested cross-validation procedure, for most
practical purposes.  Also there is some indication that the
conclusions remain even for data sets larger than the ones tested.

There is also a strong evidence that in any selection process,
regardless of the algorithms that are being selected, provided they
all have a low number of hyperparameters, one can use the flat
cross-validation procedure to simultaneously select the algorithm and
the hyperparameters, and again for all practical purposes, that
algorithm would perform as well as the algorithm selected using nested
cross-validation.


\appendices

\section{Other analysis methods and irrelevance threshold} \label{othermet}

As discussed, the analysis method in this paper assumes that each repetition is an independent experiment, and the repetitions are only aggregated at the last step, to compute the accuracy gain and the threshold of  irrelevance for a data set.   
But there are some alternatives to that analysis method. The first alternative is to consider the repetition as a way of obtaining multiple estimates for each of the accuracies measures. Thus, all measured accuracies are first averaged across the six repetition and only then used in the procedure, that is:
\begin{align*}
\est^{n}(a,i) = \frac{1}{6} \sum_r \est^{n}(a,i,r)\\ 
\est^{f}(a,i) = \frac{1}{6} \sum_r \est^{f}(a,i,r)
\end{align*}
The flat and nested selections for each data set ($f(i)$ and $n(i)$) would be selected using $\est^{n}(a,i)$ and $\est^{f}(a,i)$ (in contrast to the method used which selects 
$f(i,r)$ and $n(i,r)$ for each repetition). Then equations \ref{eq10} and \ref{eq11} would be
\begin{align*} 
\begin{split}
\ess(n,i) &= \frac{1}{6} \sum_r \acc(TE^i_r|n(i),TR^i_r,  \hat{\theta}_{ni}). \\
\ess(f,i) &= \frac{1}{6} \sum_r \acc(TE^i_r|f(i),TR^i_r,  \hat{\theta}_{fi}). 
\end{split}
\end{align*}
and 
\begin{align*} 
accgain(i) &= \ess(n,i)-\ess(f,i) 
\end{align*}
Similarly, the thresholds of irrelevance are not defined for each repetition but only for each data set:
\begin{equation*}
\delta(a,i) = | \est^n(a,i) - \ess(a,i) |
\end{equation*}

The second alternative is to consider each repetition as an independent experiment at par with the data set themselves. The results for each data set it only aggregated at the last level, when considering the p-value of the Wilcoxon test that compares $|accgain(i)|-\delta(i)$ with 0. In this second alternative, we would perform the Wilcoxon test to compare $|accgain(i,r)|-\delta(i,r)$ to 0.

Finally, there is another measure that can play the role of irrelevance threshold. When we discussed the irrelevance threshold we mention unavoidable error or variance and we chose the mean difference between the nested estimate of accuracy and the true measure of accuracy on the test set. But standard deviation is a common way of measuring error we could use it instead of the mean difference of two accuracies as we did. There are three measures of accuracy for each repetition: $\est^{n}(a,i,r)$, $\est^{f}(a,i,r)$ and  $\acc(TE^i_r|a,TR^i_r, \hat{\theta}_{ai}) $. We define the threshold for each data set as the smallest of the three standard deviations of measured accuracies across the six repetitions:
\begin{equation*}
\delta(a,i) = \min   \left\{ \begin{matrix}
\sigma_r(\est^{n}(a,i,r))\\
\sigma_r(\est^{f}(a,i,r)) \\
\sigma_r(\acc(TE^i_r|a,TR^i_r,\hat{\theta}_{ai}))
\end{matrix} \right.
\end{equation*}

The results of both the first alternative and second alternative analysis methods are reported in Table~\ref{tabapp1}. The third block of results use  minimum of the standard deviations as the threshold of irrelevance (and using the paper's original method of analysis).

\begin{table}[ht]
\caption{Results for the two alternative analysis methods and the alternative definition of the threshold of irrelevance}\label{tabapp1}
\centering
\begin{tabular}{lrrrrr}
  \hline
scenario &  same choice   & p.value& mean & low CI & high CI\\
\hline
\multicolumn{5}{c}{first alternative analysis}\\
\hline
top 3 & 78\%& 6.32e-06 & -0.005 & -0.015 &  -0.002 \\
full & 72\% & 0.0005 & -0.005 & -0.015 & -0.002 \\
\hline 
\multicolumn{5}{c}{second alternative analysis}\\
  \hline
top 3 & 78\%& 3.05e-13 & -0.002 & -0.002 &  -0.001 \\
full & 72\% & 2.67e-08 & -0.001 & -0.002 & -0.001 \\
\hline 
\multicolumn{5}{c}{other definition of irrelevance}\\
\hline
top 3 & 80\%& 5.94e-06& -0.002 & -0.003 &  -0.001 \\
full & 71\% & 1e-06& -0.002 & -0.003 & -0.001 \\
\hline
\end{tabular}
\end{table}

The two different methods of analysis and the other definition of the irrelevance threshold yield results that are consistent with the claims of the paper.

\section{Should one select the algorithm at all?}\label{oneshot}

Given that our research shows an unexpected result that flat CV is acceptable as a method to select classification algorithms, contrary to the common practice in Machine Learning, we decided to explore another unexpected result, whether the selection of algorithms is really necessary, or if one should just use random forests, which was the best ranked algorithm in the experiments. We compared the decision of using only rf against the nested procedure. The results are in Table~\ref{tabapp2}

\begin{table}[ht]
\caption{Results for only choosing random forest as the classifier against selecting the best using the nested procedure.}\label{tabapp2}
\begin{tabular}{lrrrrr}
  \hline
scenario &  same choice   & p.value & mean & low CI & high CI\\
\hline
full & 28\% & 1& 0.011 & 0.006 & 0.024 \\
  \hline
\end{tabular}
\end{table}

The results show that the accuracy gain is certainly above the threshold of irrelevance, and thus selecting the algorithm results in a expected accuracy gain of practical consequence. One cannot assume that random forest is likely result in a classifier within the irrelevance threshold of the best option, and thus avoid the search altogether. 




\end{document}